\documentclass{opt2020} % Include author names
\usepackage[fixamsmath]{mathtools}
\mathtoolsset{showonlyrefs,showmanualtags,mathic,centercolon} % number all equations

\makeatletter
\def\set@curr@file#1{\def\@curr@file{#1}} %temp workaround for 2019 latex release
\makeatother
\usepackage[load-configurations=version-1]{siunitx} % newer version
\newcommand{\scn}[2]{#1\mathrm{e}{#2}}
\newcommand{\R}{\mathbb{R}}
\renewcommand{\t}[1]{\widetilde{#1}} % tilde

\newcommand{\bigO}{\mathcal{O}}
\newcommand{\prox}{\operatorname{prox}}
\newcommand{\grad}{\nabla}
\DeclareMathOperator*{\argmin}{arg\,min} % argmin
\newcommand{\appref}[1]{Appendix~\ref{#1}}
\newcommand{\secref}[1]{Section~\ref{#1}}

\newcommand{\figref}[1]{Figure~\ref{#1}}

\title[Efficient Hyperparameter Tuning with DFO]{Efficient Hyperparameter Tuning with Dynamic Accuracy Derivative-Free Optimization}

\optauthor{%
 \Name{Matthias J.~Ehrhardt} \Email{m.ehrhardt@bath.ac.uk}\\
 \addr Institute for Mathematical Innovation and Department of Mathematical Sciences, University of Bath
 \AND
 \Name{Lindon Roberts} \Email{lindon.roberts@anu.edu.au}\\
 \addr Mathematical Sciences Institute, Australian National University%
}

\begin{document}

\maketitle

\begin{abstract}%
Many machine learning solutions are framed as optimization problems which rely on good hyperparameters.
Algorithms for tuning these hyperparameters usually assume access to exact solutions to the underlying learning problem, which is typically not practical.
Here, we apply a recent dynamic accuracy derivative-free optimization method to hyperparameter tuning, which allows inexact evaluations of the learning problem while retaining convergence guarantees.
We test the method on the problem of learning elastic net weights for a logistic classifier, and demonstrate its robustness and efficiency compared to a fixed accuracy approach.
This demonstrates a promising approach for hyperparameter tuning, with both convergence guarantees and practical performance.
\end{abstract}

%\begin{keywords}%
%  List of keywords%
%\end{keywords}

\section{Introduction}
Tuning model hyperparameters is a common problem encountered in machine learning for estimating regularization parameters, learning rates, and many other important quantities.
In many cases, hyperparameter tuning can be viewed as a bilevel optimization problem for hyperparameters $\theta$:
\begin{align}
    \min_{\theta,w} F(\theta, w) \qquad \text{subject to} \qquad w \in \argmin_{\t{w}} \Phi(\t{w}, \theta), \label{eq_bilevel}
\end{align}
where the lower-level objective $\Phi$ corresponds to a learning problem for weights $w$, and the upper-level objective $F$ typically measures a test error.
Hyperparameter tuning often uses global optimization methods \cite{Feurer2019}, but recent work has also considered local gradient-based \cite{Luketina2016,Shaban2019} and derivative-free \cite{Ghanbari2017,Lakhmiri2019gerad} methods.
The lower-level problem is usually solved with iterative methods, so only inexact evaluations of feasible $w$ are available.
As a result, few algorithms have convergence theory; exceptions include \cite{Li2018}, \cite{Shaban2019} under specific assumptions and \cite{Ghanbari2017} (in light of \cite{Chen2016}).
Bilevel learning is also used in inverse problems  (e.g.~\cite{Arridge2019,Kunisch2013bilevel,Ochs2015,Sherry2019samplingpublished}), although other approaches exist \cite{Engl1996,Benning2018actanumerica,Hansen1992lcurve}.

Recently in \cite{Ehrhardt2020}, a model-based DFO method \cite{Conn2009,Audet2017} for bilevel learning (in an inverse problems context) was proposed, which has convergence guarantees while still allowing inexact lower-level minimizers (but with a dynamic accuracy controlled by the algorithm).
Here, we apply this algorithm to bilevel optimization for hyperparameter tuning \eqref{eq_bilevel}.
Specifically, we apply the framework of \cite{Ehrhardt2020} to the case where the lower-level objective is nonsmooth (but strongly convex).
Our approach has similarities to Hyperband \cite{Li2018}, which considers an inexact Bayesian optimization method, whereas our approach is designed to exploit the bilevel structure to find a local minimum.
We can achieve the dynamic accuracy requirements for the upper-level algorithm, but delegate full convergence theory to future work (noting that model-based DFO methods for nonsmooth problems exist; e.g.~\cite{Garmanjani2016,Grapiglia2016,Khan2018}).
We show results for tuning elastic net weights of a logistic classifier.
Compared to fixed-accuracy variants of the same DFO method, our approach has strong computational performance and robustness, and avoids manually tuning the lower-level problem solution accuracy.

%\newpage

\section{Lower-Level Problem}
First, we consider a general formulation for solving nonsmooth but strongly convex learning problems.
Suppose that we wish to learn weights $w\in\R^d$ by minimizing a loss function which depends on parameters $\theta\in\R^m$; that is, we wish to solve
\begin{align}
    \hat{w}(\theta) := \argmin_{w\in\R^d} \Phi(w,\theta) := f(w,\theta) + g(w,\theta), \label{eq_lower_level_generic}
\end{align}
where $f(\cdot,\theta)$ is $\mu(\theta)$-strongly convex and smooth with $L(\theta)$-Lipschitz continuous gradient, and $g(\cdot,\theta)$ is convex and possibly nonsmooth.
As a consequence, we note that $\Phi(w,\theta)$ is also $\mu(\theta)$-strongly convex \cite[Lemma 5.20]{Beck2017}.
This is a standard problem type, and in this work we solve \eqref{eq_lower_level_generic} using a variant of FISTA \cite{Beck2009} designed for strongly convex objectives \cite[Algorithm 5]{Chambolle2016actanumerica}:
\begin{align}\begin{aligned}
t_{k+1} &= \left[1 - q t_k^2 + \sqrt{(1 - q t_k^2)^2 + 4 t_k^2}\right]/2, &
\beta_{k+1} &= \left[(t_k - 1)(1 - t_{k+1} q)\right] / \left[t_{k+1}(1 - q)\right], \\
z^{k+1} &= w^k + \beta_{k+1}(w^k - w^{k-1}), &
w^{k+1} &= \prox_{\tau g(\cdot, \theta)} (z^{k+1} - \tau \nabla_w f(z^{k+1}, \theta)), \\
\end{aligned}\label{EQ:FISTA}
\end{align}
where $w^{-1}:=w^0$, $q:=\tau \mu(\theta)$, and we choose $\tau=1/L(\theta)$ and $t_0=0$.
This algorithm achieves linear convergence to $\hat{w}(\theta)$ with the a priori estimate (from \cite[Theorem 4.10]{Chambolle2016actanumerica} and $\Phi(w^k,\theta)-\Phi(\hat{w}(\theta),\theta)\geq (\mu/2)\|w^k-\hat{w}(\theta)\|_2^2$ using strong convexity, and defining $\kappa(\theta):=L(\theta)/\mu(\theta)$):
\begin{align}
    \|w^k - \hat{w}(\theta)\|_2^2 \leq \left(1 - \kappa(\theta)^{-1/2}\right)^k \left[\kappa(\theta)\left(1+\kappa(\theta)^{-1/2}\right) \|w^0 -\hat{w}(\theta)\|_2^2\right] \, , \label{eq_fista_rate}
\end{align}

\paragraph{Guaranteeing Sufficient Accuracy}
In the dynamic accuracy DFO framework \cite{Ehrhardt2020}, evaluations of $\hat{w}$ need to sufficiently accurate in that sense that we terminate the iteration \eqref{EQ:FISTA} once $\|w^k-\hat{w}(\theta)\|_2^2 \leq \epsilon$
is achieved, for some $\epsilon>0$ specified by the upper-level solver.
A priori, we can determine the required number of iterations using \eqref{eq_fista_rate}.
However, since it is difficult to estimate $\|w^0 -\hat{w}(\theta)\|_2$ in practice, we also consider the a posteriori termination criterion
\begin{align}
    \|w^k-\hat{w}(\theta)\|_2^2 \leq \|d(w^k)\|_2^2 / \mu(\theta)^2 \leq \epsilon, \label{eq_a_posteriori_bound}
\end{align}
where $d(w^k)\in\partial \Phi(w^k,\theta)$, which ensures the required accuracy from the $\mu(\theta)$-strong convexity of $\Phi(\cdot,\theta)$ \cite[Theorem 5.24]{Beck2017}.
In FISTA, we calculate $d(w^k)$ as
\begin{align}
    d(w^k) = \grad_w f(w^{k+1}, \theta) - \grad_w f(z^{k+1}, \theta) + \frac{1}{\tau}(z^{k+1}-w^{k+1}) \in \partial \Phi(w^{k+1},\theta),
\end{align}
which follows from noting $z^{k+1}-\tau\grad_w f(z^{k+1},\theta)-w^{k+1}\in\partial (\tau g(\cdot,\theta))$, which follows from the properties of the proximal operator \cite[Theorem 6.39]{Beck2017}.

\begin{figure}[t]
    \centering
    \includegraphics[height=3cm]{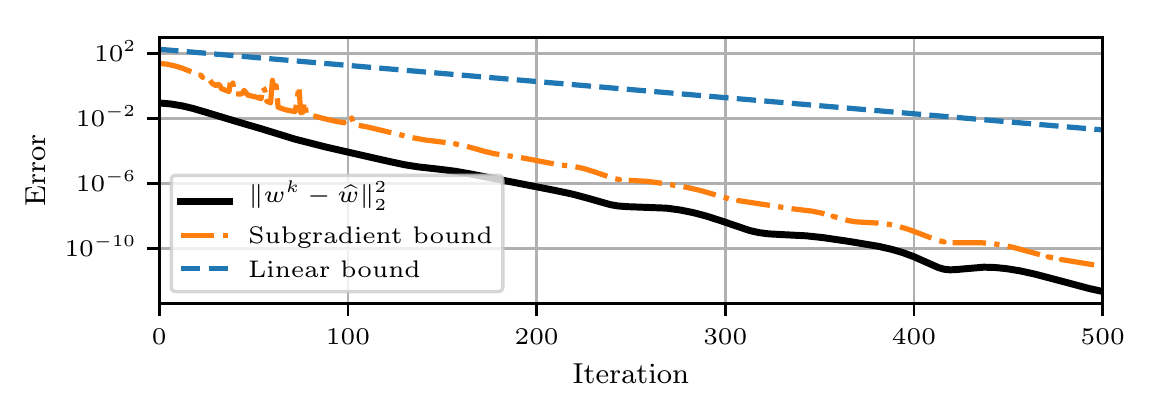}
    \caption{Comparison of the a-priori linear convergence bound \eqref{eq_fista_rate} against the a-posteriori subgradient bound \eqref{eq_a_posteriori_bound}. Showing results over 500 iterations FISTA for the problem \eqref{eq_lasso}.}
	\label{fig_accuracy_bounds}
\end{figure}

In \figref{fig_accuracy_bounds}, we compare the a priori \eqref{eq_fista_rate} and a posteriori error \eqref{eq_a_posteriori_bound} bounds on $\|w^k-\hat{w}(\theta)\|_2^2$ on the linear inverse problem
\begin{align}
    \Phi(w,\theta) = \frac{1}{2}\|Aw-b\|_2^2 + \frac{\theta_1}{2}\|w\|_2^2 + \theta_2\|w\|_1, \label{eq_lasso}
\end{align}
for $\theta=[\theta_1,\theta_2]^T$, where we use a randomly generated $A\in\R^{100\times 200}$, $w^0\in\R^{200}$ and $b\in\R^{100}$, and parameters $\theta=[10,10]^T$.
For this problem, we have $L(\theta)\approx\scn{1.96}{4}$ and $\mu(\theta)=10$.
Despite the a priori bound using the correct value of $\|w^0 -\hat{w}(\theta)\|_2$, the a posteriori subgradient bound gives a much tighter estimate of $\|w^k-\hat{w}(\theta)\|_2^2$.
Because of this and not being able to estimate $\|w^0 -\hat{w}(\theta)\|_2$ accurately in practice, in our numerical results we only use the subgradient bound to ensure the required accuracy.

\section{Upper-Level Problem} \label{sec_upper_level}
We now describe our algorithm for solving the upper-level problem of learning $\theta$.
Suppose we have access to $n$ lower-level problems which depend on the same parameters $\theta\in\R^m$, and that each lower-level minimizer $\hat{w}_j(\theta)\in\R^{d_j}$ has an associated loss function $\ell_j:\R^{d_j}\to\R$; in \secref{sec_numerics}, this will be a measure of test error.
Then, our upper-level problem is
\begin{align}
    \min_{\theta\in\R^m} &\:\: F(\theta) := \sum_{j=1}^{n} \ell_j(\hat{w}_j(\theta)) + \mathcal{J}(\theta), \quad \text{subject to} \quad \hat{w}_j(\theta) := \argmin_{w\in\R^d} \Phi_j(w, \theta), \label{eq_upper_level}
\end{align}
where $\mathcal{J}(\theta)$ is an optional regularization term for $\theta$, and $\Phi_j$ is strongly convex in $w$ (but possibly nonsmooth) and smooth in $\theta$. 
Under these assumptions, there are various results showing that $\hat{w}_j(\theta)$ is locally Lipschitz in $\theta$ for a variety of problems; e.g.~\cite{Hintermuller2015,Riis2018}.

To solve \eqref{eq_upper_level} we follow \cite{Ehrhardt2020} and use a model-based DFO method.
Here, at every iteration $k$ we have a collection of points $\{y^{(k)}_0,\ldots,y^{(k)}_m\}$ at which we have evaluated $\hat{w}_j$ (for each $j$), to some accuracy.
We use this information to build a local quadratic model $m_k(\theta)$ which approximates $F(\theta)$ near $\theta^{(k)}$ by requiring that the model interpolate $F$ at each of $y^{(k)}_0,\ldots,y^{(k)}_m$.
Our upper-level iteration can be summarized as:
\begin{enumerate}
    \item Build the interpolating local quadratic model $m_k(\theta)\approx F(\theta)$ using (inexact) evaluations of $\hat{w}_j(y^{(k)}_0),\ldots,\hat{w}_j(y^{(k)}_m)$.
    \item Calculate a minimizer $s^{(k)}$ of the model inside a trust region, namely solve (possibly inexactly) $\min_{\|s\|_2\leq\Delta_k} m_k(\theta^{(k)}+s)$,
    for a given trust-region radius $\Delta_k>0$.
    \item Ensure $F(\theta^{(k)})$ and $F(\theta^{(k)}+s^{(k)})$ are evaluated to sufficiently high accuracy, and then if sufficient decrease is achieved, set $\theta^{(k+1)}=\theta^{(k)}+s^{(k)}$ and increase $\Delta_k$, otherwise set $\theta^{(k+1)}=\theta^{(k)}$ and decrease $\Delta_k$.
    \item Replace one interpolation point with $\theta^{(k)}+s^{(k)}$.
\end{enumerate}
This algorithm typically requires that our inexact lower-level evaluations $w_j(\theta)\approx \hat{w}_j(\theta)$ satisfy $\|w_j(\theta)-\hat{w}_j(\theta)\|_2 \leq c\,\Delta_k^2$ for some $c>0$ (where we use $c=100$ in \secref{sec_numerics} below).

Full details of a suitable dynamic accuracy upper-level algorithm are given in \cite{Ehrhardt2020}.
There, it is shown that, if $\mathcal{J}$, and each $\ell_j$ and $\Phi_j$ are sufficiently smooth, and each $\Phi_j$ is strongly convex in $w$, then the model-based DFO algorithm finds a $\theta$ with $\|\grad F(\theta)\|_2\leq\epsilon$ in at most $\bigO(\epsilon^{-2})$ upper-level iterations, or at most $\bigO(\epsilon^{-2} |\log\epsilon|)$ lower-level iterations.

%\newpage

\section{Numerical Results} \label{sec_numerics}
\paragraph{Summary of Test Problem}
We test our approach by attempting to learn elastic net regularizer \cite{Zou2005} weights $\theta\in\R^2$ for a logistic classifier of each digit 0--5 in MNIST \cite{lecun2010mnist} (i.e.~each lower-level problem corresponds to a classifier for a different digit and the upper-level problem measures a test set error).
Below, we validate the results of our approach by considering the remaining digits.
To test our approach, we follow \cite{Ehrhardt2020} and use a version of DFO-LS \cite{Cartis2019a,Cartis2018} which is modified to handle dynamic accuracy evaluations as per \secref{sec_upper_level}.
We use FISTA \eqref{EQ:FISTA} for the lower-level solver, and compare the dynamic accuracy approach with a combination of FISTA and DFO-LS, but using a fixed number of FISTA iterations $K$ for every lower-level solve.
Full details are given in \appref{app_numerics_details}.

\paragraph{General Comparison}
\figref{fig_start0} compares the dynamic-accuracy variant against fixed-accuracy variants with low ($K=20$), medium ($K=200$) and high accuracy ($K=2000$), 
with upper-level initialization $\theta^{(0)}=[1,1]^T$ in all cases.
Overall, dynamic accuracy is comparable to medium accuracy: it reaches the same $\theta$ at a similar speed.
By comparison, low accuracy gives a quite different value of $\theta$ to the others, and high accuracy takes substantially more computational effort to converge.
A key point to note here, is that dynamic accuracy performs similar to the best fixed-accuracy variant without requiring the necessary accuracy a priori.

%\begin{figure}[t]
%    \centering
%    \subfloat[Objective $F(\theta)$]{\label{fig_start0_obj}\includegraphics[height=3cm,width=3.6cm]{img/elastic_net_start_0_reduced_list_obj_redn2.pdf}}
%    ~
%    \subfloat[Total FISTA iterations]{\label{fig_start0_niters}\includegraphics[height=3cm,width=3.6cm]{img/elastic_net_start_0_reduced_list_niters_cumulative.pdf}}
%    ~
%    \subfloat[Best $\ell_1$ penalty $10^{\theta_1}$]{\label{fig_start0_param1}\includegraphics[height=3cm,width=3.6cm]{img/elastic_net_start_0_reduced_list_param0.pdf}}
%    ~
%    \subfloat[Best $\ell_2$ penalty $10^{\theta_2}$]{\label{fig_start0_param2}\includegraphics[height=3cm,width=3.6cm]{img/elastic_net_start_0_reduced_list_param1.pdf}}
%    \caption{Initial results using $\theta^{(0)}=[1,1]^T$.}
%	\label{fig_start0}
%\end{figure}

\begin{figure}[t]  
    \floatconts  
  {fig_start0}% label for whole figure  
  {\vspace{-1.5em}\caption{Initial results using $\theta^{(0)}=[1,1]^T$.}}% caption for whole figure  
  {%  
    \subfigure[Objective $F(\theta)$]{%  
      \label{fig_start0_obj}% label for this sub-figure  
      \includegraphics[height=3cm,width=3.6cm]{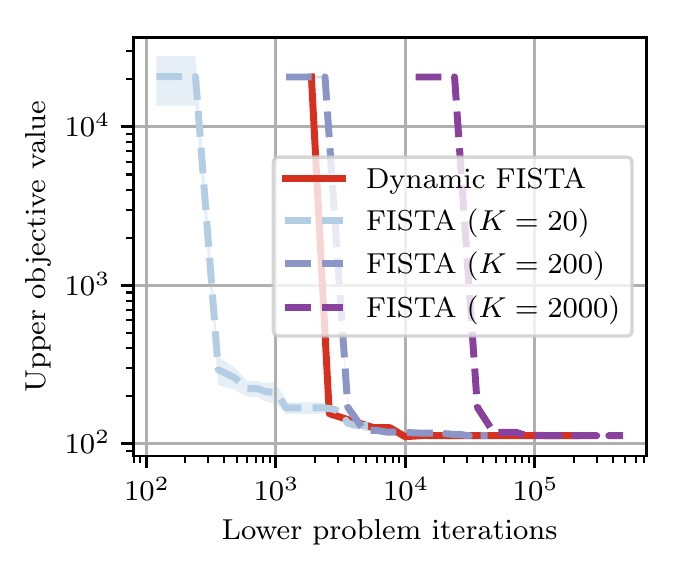}  
    }
    \hfill % space out the images a bit  
    \subfigure[FISTA iterations]{%  
      \label{fig_start0_niters}% label for this sub-figure  
      \includegraphics[height=3cm,width=3.6cm]{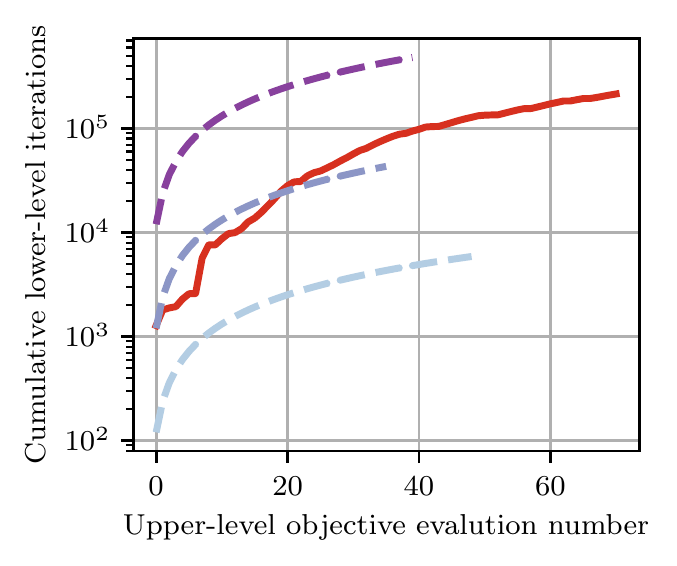}
    }
    ~ % space out the images a bit  
    \subfigure[Best $\ell_1$ penalty $10^{\theta_1}$]{%  
      \label{fig_start0_param1}% label for this sub-figure  
      \includegraphics[height=3cm,width=3.6cm]{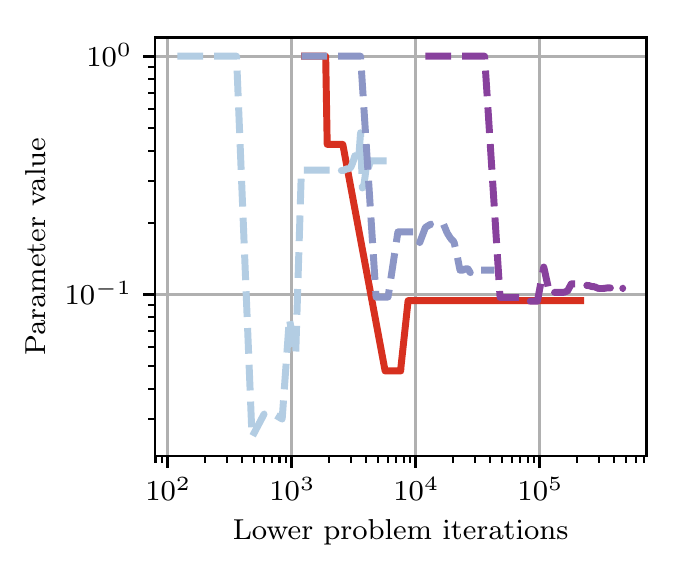}
    }
    \hfill % space out the images a bit  
    \subfigure[Best $\ell_2$ penalty $10^{\theta_2}$]{%  
      \label{fig_start0_param2}% label for this sub-figure  
      \includegraphics[height=3cm,width=3.6cm]{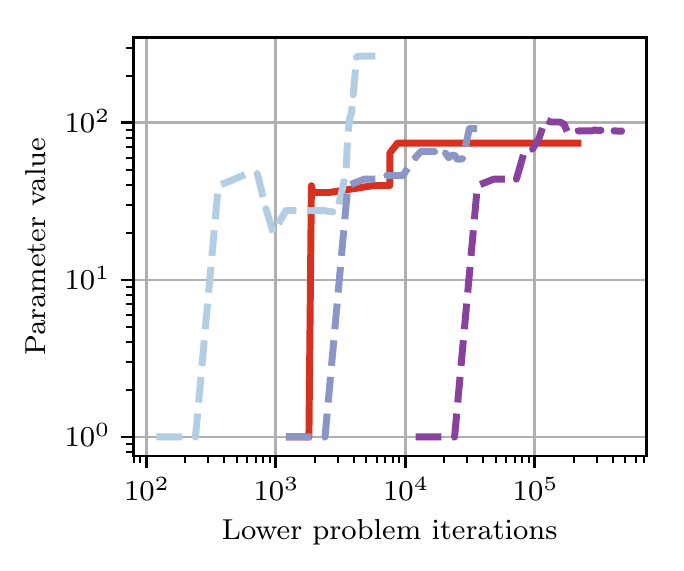}
    }
  }  
\end{figure}

\begin{figure}[t]
    \centering
    \includegraphics[height=3cm,width=15cm]{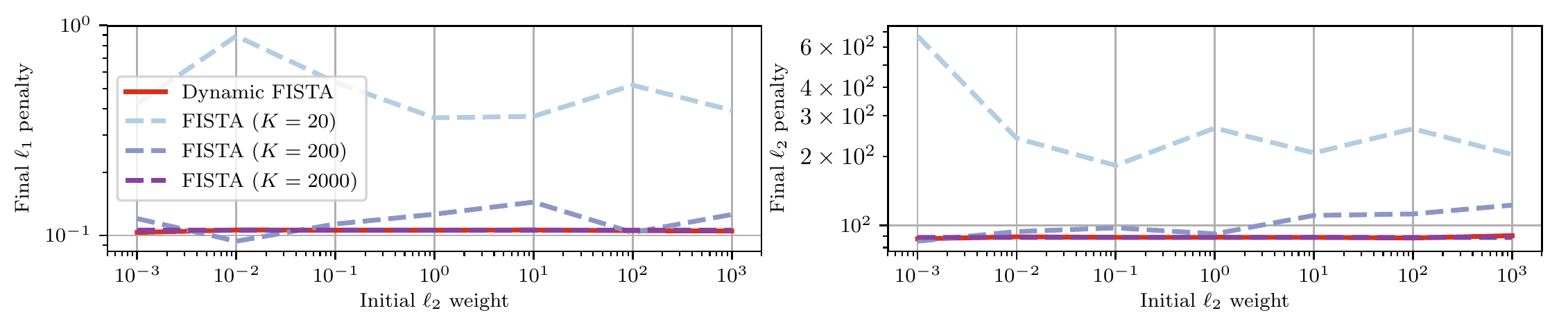}
    \caption{Final values of $\theta_1$ (left) and $\theta_2$ (right) reached for different starting points $\theta^{(0)}_2$.}
	\label{fig_starting_vary_param_compare}
\end{figure}

\begin{figure}[t]
    \centering
    \includegraphics[height=3cm,width=15cm]{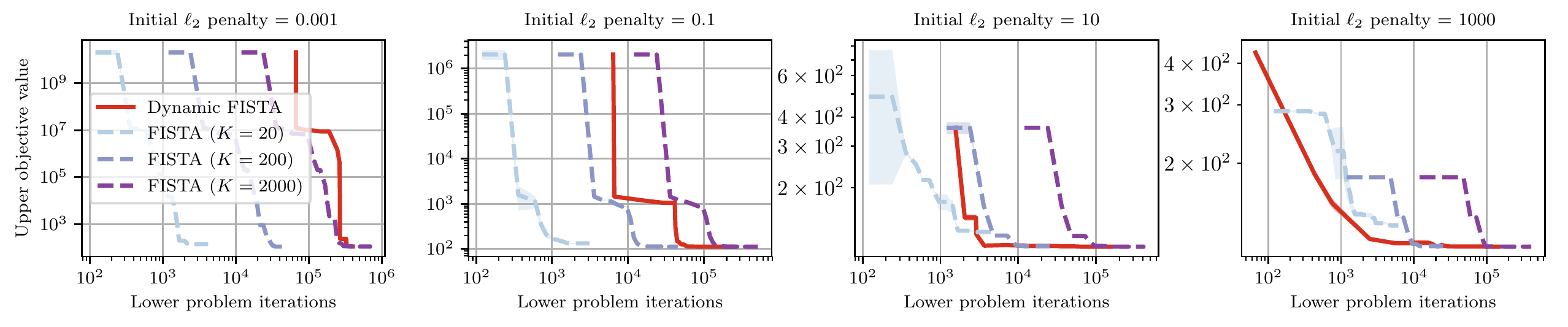}
    \caption{Comparison of objective reduction for different choices of starting point $\theta^{(0)}_2$.}
	\label{fig_starting_vary_obj}
\end{figure}

\paragraph{Robustness}
It was observed in \cite{Ehrhardt2020} that dynamic accuracy is more robust to initialization and a good initialization can accelerate its initial objective decease. 
To test these features here, we perform the same tests as above, but vary the initial $\ell_2$ penalty by choosing $\theta^{(0)}_2\in\{-3,-2,\ldots,3\}$.
\figref{fig_starting_vary_param_compare} shows that only high accuracy and dynamic accuracy give stable results, whereas the low and medium accuracy converge to different $\theta$ depending on the starting point.
\figref{fig_starting_vary_obj} shows that as $\theta^{(0)}_2$ increases,
the relative performance of dynamic accuracy improves. 
Thus, since dynamic accuracy is stable with respect to initialization, selecting $\theta^{(0)}$ to give well-conditioned lower-level problems is appropriate and the associated reduction in computational work can be achieved.

\begin{figure}[tbh]
    \centering
    \includegraphics[height=3.5cm,width=15cm]{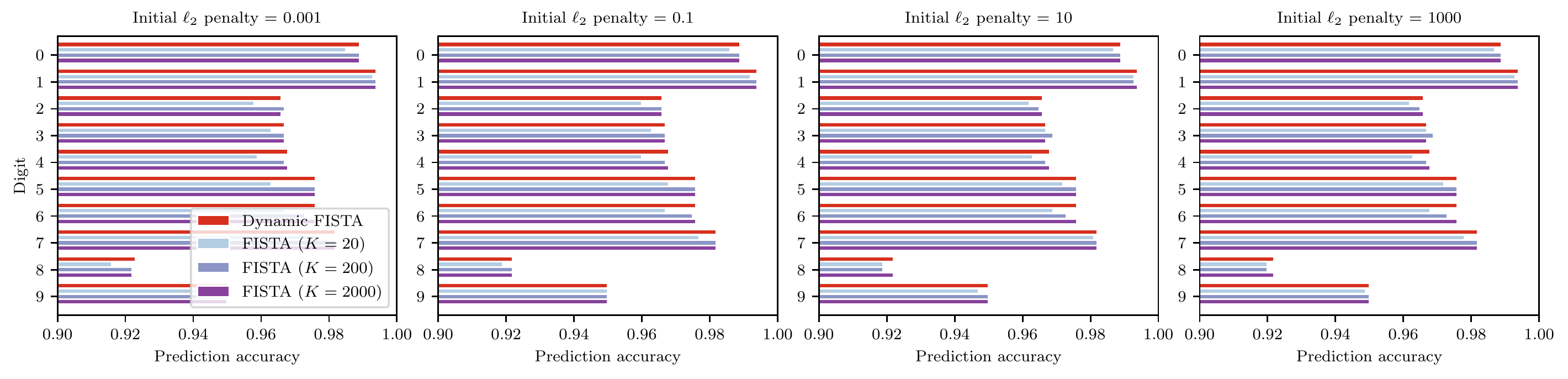}
    \caption{Comparison of validation results for different choices of starting point $\theta^{(0)}_2$.}
	\label{fig_starting_vary_validation}
\end{figure}

\paragraph{Validation}
Finally, we measure the generalizability of the upper-level minimizers.
Here, the learned values of $\theta$ from each accuracy variant and each initialization $\theta^{(0)}$ are used to train logistic classifiers for MNIST (using a new set of 5000 training images), and measure their test error (on another new set of 1000 test images).
Despite the upper-level problem only considering digits 0--5, here we look at the performance of each $\theta$ in training classifiers for all digits.
We measure the test error by looking at the proportion of test images classified correctly (using the predictor $\hat{y}^{(j)}_i=1$ if $p^{(j)}_i(\hat{w}_j(\theta))\geq 0.5$; see \appref{app_numerics_details}).
\figref{fig_starting_vary_validation} shows the validation for each digit $j\in\{0,\ldots,9\}$ and initialization $\theta^{(0)}_2$.
The overall accuracy is generally lower for digits 6--9 than the digits used in the upper-level objective (0--5).
However, we see consistently that dynamic and high accuracy achieve better prediction accuracy than low and medium accuracy.

\section{Conclusions and Further Work}
We have extended the approach in \cite{Ehrhardt2020} to the case of nonsmooth lower-level problems---we can guarantee solutions to the lower-level problem are sufficiently accurate for the upper-level algorithm.
When applied to learning elastic net regularizer weights for MNIST, we find that the dynamic accuracy DFO approach gives a robust method for hyperparameter tuning.
Compared to the fixed-accuracy variants, which have strong dependency on the selected number of iterations (both in computational speed and quality of minimizer), the dynamic-accuracy version can achieve both fast progress and high-quality minimizers, without needing to be manually tuned.
We delegate to future work a full study of the convergence of the upper-level algorithm in this setting, and comparisons between the dynamic accuracy DFO approach and other methods for hyperparameter tuning.

\subsection{Acknowledgements}
MJE acknowledges support from the EPSRC (EP/S026045/1, EP/T026693/1), the Faraday Institution (EP/T007745/1) and the Leverhulme Trust (ECF-2019-478).

%\newpage

\bibliography{refs}
%\newpage
%\clearpage
\appendix

\section{Details of Numerical Test Model} \label{app_numerics_details}
\paragraph{General Problem Formulation}
In this work, we focus on a specific choice of problem, namely selecting regularizer weights for logistic regression \cite[Chapter 8]{Murphy2012}.
Specifically, for upper-level problem $j\in\{1,\ldots,n\}$, suppose we have training pairs $\{(x^{(j)}_i,y^{(j)}_i)\}_{i=1}^{N_j}$ with features $x^{(j)}_i\in\R^{d_j}$ and binary labels $y^{(j)}_i\in\{-1,1\}$.
Minimizing the negative log-likelihood of the corresponding logistic classifier, with an elastic net regularizer \cite{Zou2005}, corresponds to the lower-level problem
\begin{align}
    \hat{w}_j(\theta) := \argmin_{w\in\R^{d_j}} \Phi_j(w,\theta) := \frac{1}{N_j}\sum_{i=1}^{N_j} \log(1 + \exp(-y^{(j)}_i w^T x^{(j)}_i)) + \frac{10^{\theta_1}}{2}\|w\|_2^2 + 10^{\theta_2} \|w\|_1, \quad \label{eq_elastic_net_logistic}
\end{align}
for (undetermined) regularizer log-weights $\theta_1,\theta_2\in\R$.
That is, we form $\Phi_j(w,\theta)$ \eqref{eq_lower_level_generic} by taking
\begin{align}
    f_j(w,\theta) := \frac{1}{N_j}\sum_{i=1}^{N_j} \log(1 + \exp(-y^{(j)}_i w^T x^{(j)}_i)) + \frac{10^{\theta_1}}{2}\|w\|_2^2, \quad \text{and} \quad g_j(w,\theta) := 10^{\theta_2} \|w\|_1,
\end{align}
and so $f_j(\cdot,\theta)$ is $\mu_j(\theta)$-strongly convex and $L_j(\theta)$-smooth with
\begin{align}
    \mu_j(\theta) := 10^{\theta_1}, \quad \text{and} \quad L_j(\theta) := \frac{1}{4N_j}\|X^{(j)}\|_2^2 + 10^{\theta_1},
\end{align}
where $X^{(j)}\in\R^{{N_j}\times {d_j}}$ has columns $x^{(j)}_i$.

As our upper-level objective, suppose for each lower-level problem $j$ we have learned weights $w_j(\theta)\in\R^{d_j}$ (in practice $w_j(\theta)$ is our approximation to $\hat{w}_j(\theta)$ from solving \eqref{eq_elastic_net_logistic} to some finite accuracy).
We then suppose we have a test set $\{(\t{x}^{(j)}_i,\t{y}^{(j)}_i)\}_{i=1}^{\t{N}_j}$, and we calculate the estimated probabilities $p^{(j)}_i(\hat{w}_j(\theta)) = \operatorname{sigm}(\hat{w}_j(\theta)^T \t{x}^{(j)}_i)$ is the probability that $\t{y}^{(j)}_i=1$ as determined by the classifier.
We then take our loss for the lower-level problem $j$ to be a smoothed approximation of the test accuracy:
\begin{align}
    \ell_j(\hat{w}_j(\theta)) := \sum_{i=1}^{\t{N}_j} \left[p^{(j)}_i(\hat{w}_j(\theta)) - \t{p}^{(j)}_i\right]^2, \qquad \text{where} \qquad \t{p}^{(j)}_i = \begin{cases}1, & \t{y}^{(j)}_i=1, \\ 0, & \text{otherwise}.\end{cases}
\end{align}
That is, $\ell_j(\theta)$ is the squared Euclidean distance from the vector $[p^{(j)}_1,\ldots,p^{(j)}_{\t{N}_j}]^T$ to the `perfect' classifier's probabilities $[\t{p}^{(j)}_1,\ldots,\t{p}^{(j)}_{\t{N}_j}]^T$.

\paragraph{Example Problem}
We take our training data from MNIST \cite{lecun2010mnist}, where we attempt to learn a classifier for each digit separately (i.e.~for digit $j\in\{0,\ldots,9\}$, learn a classifier for whether image $i$ is of digit $j$ or not).
In this case, we have the same features for all lower-level problems, and we choose $N_j=5000$ and $\t{N}_j=1000$ images for the training and test sets respectively.
To test the generalizability of our approach, we only solve the upper-level problem for digits 0--5 (i.e.~$n=6$).
We validate the final learned $\theta$ on the same problem for the remaining digits 6--9.

We augment our upper-level problem with a regularizer which encourages well-conditioned lower-level problems and a large value of $\theta_2$ (which should yield sparse weights $w_j$):
\begin{align}
    \mathcal{J}(\theta) := \alpha_1 \left(\frac{L(\theta)}{\mu(\theta)}\right)^2 + \alpha_2 10^{-\theta_2},
\end{align}
for weights $\alpha_1=10^{-8}$ and $\alpha_2=1$.

To test our approach, we follow \cite{Ehrhardt2020} and use a version of DFO-LS \cite{Cartis2019a,Cartis2018} which is modified to handle dynamic accuracy evaluations as per \secref{sec_upper_level}.
We use FISTA \eqref{EQ:FISTA} for the lower-level solver, and compare the dynamic accuracy approach with a combination of FISTA and DFO-LS, but using a fixed number of FISTA iterations $K$ for every lower-level solve.
We impose bounds $\theta\in[10^{-8}, 10^8]^2$, and terminate after 80 upper-level evaluations or the trust-region radius in DFO-LS reaches $10^{-5}$.

\end{document}